%% file: main.tex
\documentclass[10pt,twocolumn,letterpaper]{article}

\usepackage[pagenumbers]{iccv} 

\definecolor{iccvblue}{rgb}{0.21,0.49,0.74}
\usepackage[pagebackref,breaklinks,colorlinks,allcolors=iccvblue]{hyperref}
\usepackage{multirow}

\def\paperID{*****} 
\def\confName{ICCV}
\def\confYear{2025}

\title{PointVLA: Injecting the 3D World into Vision-Language-Action Models}

\begin{document}

\author{
Chengmeng Li$^{1,2,*}$\quad
Yichen Zhu$^{1,*,\dagger}$\quad
Junjie Wen$^{3}$\quad
Yan Peng$^{2}$\quad
Yaxin Peng$^{2,\dagger}$ \quad
Feifei Feng$^{1}$ \quad
\vspace{0.03in}
\\
$^1$Midea Group \quad $^2$Shanghai University \quad $^3$East China Normal University \quad  \\
\quad$\dagger$Corresponding author\vspace{0.1in} \\
\href{https://pointvla.github.io}{\color{blue}\textbf{pointvla.github.io}\xspace}\vspace{-0.3in}
}

\makeatletter
\let\@oldmaketitle\@maketitle%
\renewcommand{\@maketitle}{\@oldmaketitle
    \begin{center}
        \captionsetup{type=figure}
        \centering
        \includegraphics[width=1.0\textwidth]{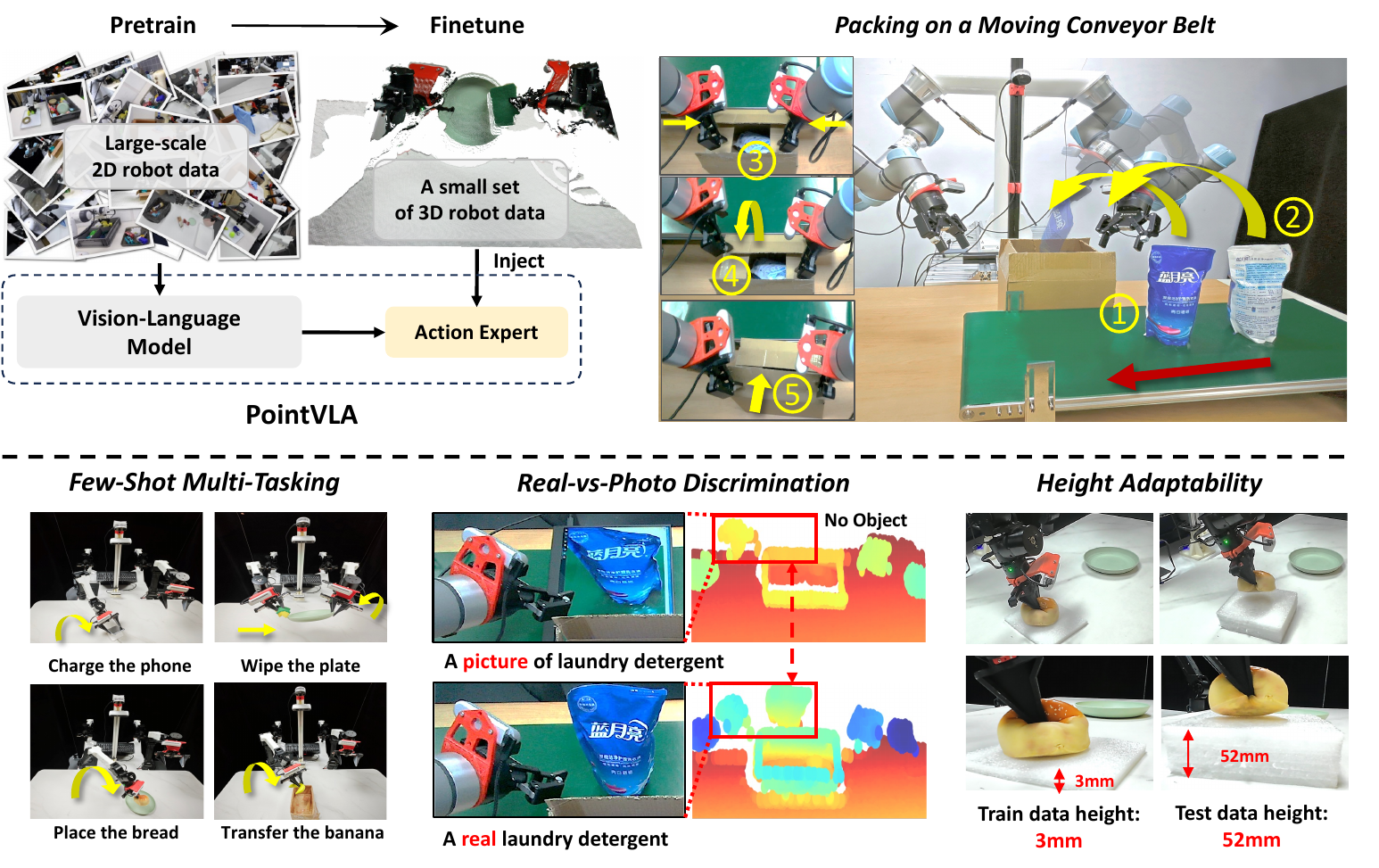}
        \caption{PointVLA builds upon the strengths of VLA, which is pre-trained on large-scale 2D robot data while incorporating 3D world into the action expert. We evaluate our method across various settings, including a challenging long-horizon packing task (top right), a few-shot multi-task setup (bottom left), and tests of the model's real-vs-photo discrimination and height adaptability.}\label{fig:framework}
    \end{center}
}
\makeatother

\maketitle

\begin{abstract}
Vision-Language-Action (VLA) models excel at robotic tasks by leveraging large-scale 2D vision-language pretraining, but their reliance on RGB images limits spatial reasoning critical for real-world interaction. Retraining these models with 3D data is computationally prohibitive, while discarding existing 2D datasets wastes valuable resources. To bridge this gap, we propose PointVLA, a framework that enhances pre-trained VLAs with point cloud inputs without requiring retraining. Our method freezes the vanilla action expert and injects 3D features via a lightweight modular block. To identify the most effective way of integrating point cloud representations, we conduct a skip-block analysis to pinpoint less useful blocks in the vanilla action expert, ensuring that 3D features are injected only into these blocks—minimizing disruption to pre-trained representations.

Extensive experiments demonstrate that PointVLA outperforms state-of-the-art 2D imitation learning methods, such as OpenVLA~\cite{openvla}, Diffusion Policy~\cite{diffusion-policy} and DexVLA~\cite{wen2025dexvla}, across both simulated and real-world robotic tasks. Specifically, we highlight several key advantages of PointVLA enabled by point cloud integration: (1) \textbf{Few-shot multi-tasking}, where PointVLA successfully performs four different tasks using only 20 demonstrations each; (2) \textbf{Real-vs-photo discrimination}, where PointVLA distinguishes real objects from their images, leveraging 3D world knowledge to improve safety and reliability; (3) \textbf{Height adaptability}, Unlike conventional 2D imitation learning methods, PointVLA enables robots to adapt to objects at varying table height that unseen in train data. Furthermore, PointVLA achieves strong performance in long-horizon tasks, such as picking and packing objects from a moving conveyor belt, showcasing its ability to generalize across complex, dynamic environments.
\end{abstract}

\section{Introduction}
\label{sec:intro}
Robot foundation models, particularly Vision-Language-Action (VLA) models~\cite{[pi0,openvla,rt-2,wen2024tinyvla,wen2025dexvla}, have demonstrated remarkable capabilities in enabling robots to perceive, understand, and interact with the physical world. These models leverage pre-trained vision-language models (VLMs)~\cite{chen2024internvl,karamcheti2024prismatic,wang2024qwen2,lu2024deepseek-vl,beyer2024paligemma} as backbones for processing visual and linguistic information, embedding them into a shared representation space, and subsequently translating them into robot actions. This process allows robots to interact with their environment in a meaningful way. The strength of a VLA model is largely dependent on the scale and quality of its training data. For instance, OpenVLA~\cite{openvla} is trained on 4k hours of open-source datasets, whereas more advanced models like $\pi_{0}$ utilize 10k hours of proprietary data, leading to significantly improved performance. In addition to these large-scale foundation models, numerous projects have contributed extensive datasets collected from real-world human demonstrations on physical robots. For example, AgiBot-World~\cite{bu2025agibotworld} has released a substantial dataset containing millions of trajectories demonstrating complex humanoid interactions. These pre-trained VLA models, along with open-source robotic datasets, have significantly advanced robot learning by providing a wealth of diverse and high-quality training data.

Despite these advances, most existing robot foundation models~\cite{rt-2,[pi0,openvla,kaushik2020fast,wen2025dexvla} are trained on 2D visual inputs~\cite{o2023open-x,khazatsky2024droid}. This represents a crucial limitation because humans perceive and interact with the world in three dimensions. The lack of comprehensive 3D spatial information in training data hinders a robot's ability to develop a deep understanding of its environment. This is particularly critical for tasks that demand precise spatial awareness, depth perception, and object manipulation. What motivates us is the fact that many organizations have already invested heavily in foundational VLA models and large-scale 2D robot datasets. Retraining these models from scratch with 3D data would be computationally prohibitive, and discarding valuable 2D robot data is impractical. Therefore, it is essential to explore novel frameworks that can integrate additional 3D input into pre-existing foundation robot models, a research area that has not been underexplored in the previous literature.

In this paper, we introduce PointVLA, a novel framework that integrates point clouds into pre-trained vision-language-action models. We assume that the new 3D robot data is significantly smaller than the pre-trained 2D data. In this scenario, it is critical not to disrupt the well-established 2D feature representations. To address this, we propose a 3D modular block that injects point cloud information directly into the action expert. By keeping the vision-language backbone intact, we ensure that the 2D visual-text embeddings are preserved and remain a reliable source of information. Furthermore, we aim to minimize disruptions to the action expert’s feature space. Through a skip block analysis, we identify layers that are less critical at test time. The feature embeddings of these ''less useful'' layers are more adaptable to the new modality. After identifying these less essential blocks, we inject the extracted 3D features via an additive approach. Our overall method maintains the integrity of the pre-trained VLA while incorporating the advantages of point cloud inputs.

We conduct extensive experiments to validate the effectiveness of our approach. For instance, on RoboTwin~\cite{mu2024robotwin} simulation platform, our method outperforms purely 3D imitation learning methods like the 3D Diffusion Policy~\cite{ze20243d}. Additionally, we conduct real-world experiments on two types of bimanual robots: humanoid-like UR5e arms and AglieX arms resembling the Aloha platform. Furthermore, our experiments highlight several key advantages of PointVLA:

\begin{itemize}
    \item \textbf{Few-shot multi-tasking}: PointVLA can perform four tasks following instruction, with each task trained on only 20 demonstrations. Training on such a small dataset across multiple tasks is challenging, and our approach significantly outperforms the baseline.
    \item \textbf{Real-vs-photo discrimination}: Real objects and their photographic representations can appear very similar in 2D images, potentially leading to confusion and safety hazards for robots. PointVLA can reliably differentiate between real objects and their images, avoiding deception by unreal objects.
    \item \textbf{Height adaptability}: PointVLA can adjust robot actions in response to changes in table height (e.g., grasping the same item on a much higher table), a scenario where traditional 2D VLA models typically fail.

\end{itemize}

Furthermore, we tackle challenging, long-horizon tasks such as picking multiple items from a moving conveyor belt and packing them into a box. These experiments demonstrate the robust performance and generalizability of our proposed PointVLA framework across diverse scenarios, suggesting a promising direction for integrating additional modalities into pre-trained VLA models.

\section{Related Works}

\textbf{Vision-Language-Action models.} Recent research has increasingly focused on developing generalist robot policies trained on large-scale robotic learning datasets~\cite{o2023open-x, khazatsky2024droid, fang2023rh20t, dasari2024ditpolicy, lin2024datascalinglawsimitation}. Vision-Language-Action (VLA) models have emerged as a promising approach for training such policies~\cite{kim24openvla, embodiedcot, [pi0, wen2025dexvla, pertsch2025fast, niu2024llarva, diffusion-policy, zhou2025chatvla, ding2025humanoid, zhao2025vlas, ding2024quar, tong2024quart, wen2024tinyvla}. VLAs extend vision-language models (VLMs)—pre-trained on massive internet-scale image and text datasets~\cite{zhu2024mipha, zhao2024cobra, zhu2024llavaphi, karamcheti2024prismatic, wang2024qwen2, lu2024deepseek-vl, llava, llava1.5, abdin2024phi3, chen2024internvl}—to robotic control~\cite{wang2024distrl}. This approach provides several key advantages: leveraging large-scale vision-language model backbones with billions of parameters enables effective learning from vast robotic datasets, while reusing pre-trained weights from internet-scale data enhances VLAs’ ability to interpret diverse language commands and generalize to novel objects and environments, making them highly adaptable for real-world robotic applications. 

\textbf{Robot learning with 3D modalities.} Learning robust visuomotor policy in 3D scene~\cite{3d_diffuser_actor, leo3d, qu2025spatialvla, ze20243d, ze2024generalizable, wang2024rise,zhang2024dcpi,goyal2023rvt,goyal2024rvt2,singh2023selfsupervised,chen2024sugar,jia2024lift3d,ze2023gnfactor} is an important domain in robot learning. Existing approaches like 3DVLA~\cite{leo3d} have proposed comprehensive frameworks that integrate diverse 3D tasks, such as generalization, visual question answering (VQA), 3D scene understanding, and robot control, into unified vision-language-action models. However, a limitation of 3DVLA is its reliance on simulation for robot control experiments, which presents a significant sim-to-real gap. Other works, such as 3D diffusion policies~\cite{ze20243d}, have demonstrated that using extrinsic 3D input (e.g., from external cameras) can improve model generalization to varying lighting conditions and object attributes. iDP3~\cite{ze2024generalizable} further enhanced the 3D visual encoder and applied it to humanoid robots, achieving robust performance in diverse environments with both egocentric and external camera views. However, discarding existing 2D robot data or completely retraining the foundation model with added 3D visual input would be computationally expensive and resource-intensive. A more practical solution is to develop a method that integrates 3D visual input as a supplementary knowledge source into a well pre-trained foundation model, thereby reaping the benefits of new modalities without compromising the performance of the trained model.

\section{Methodology}
\subsection{Preliminaries: Vision-Language-Action Models}
\begin{figure*}[t]
    \centering
    \includegraphics[width=0.9\textwidth]{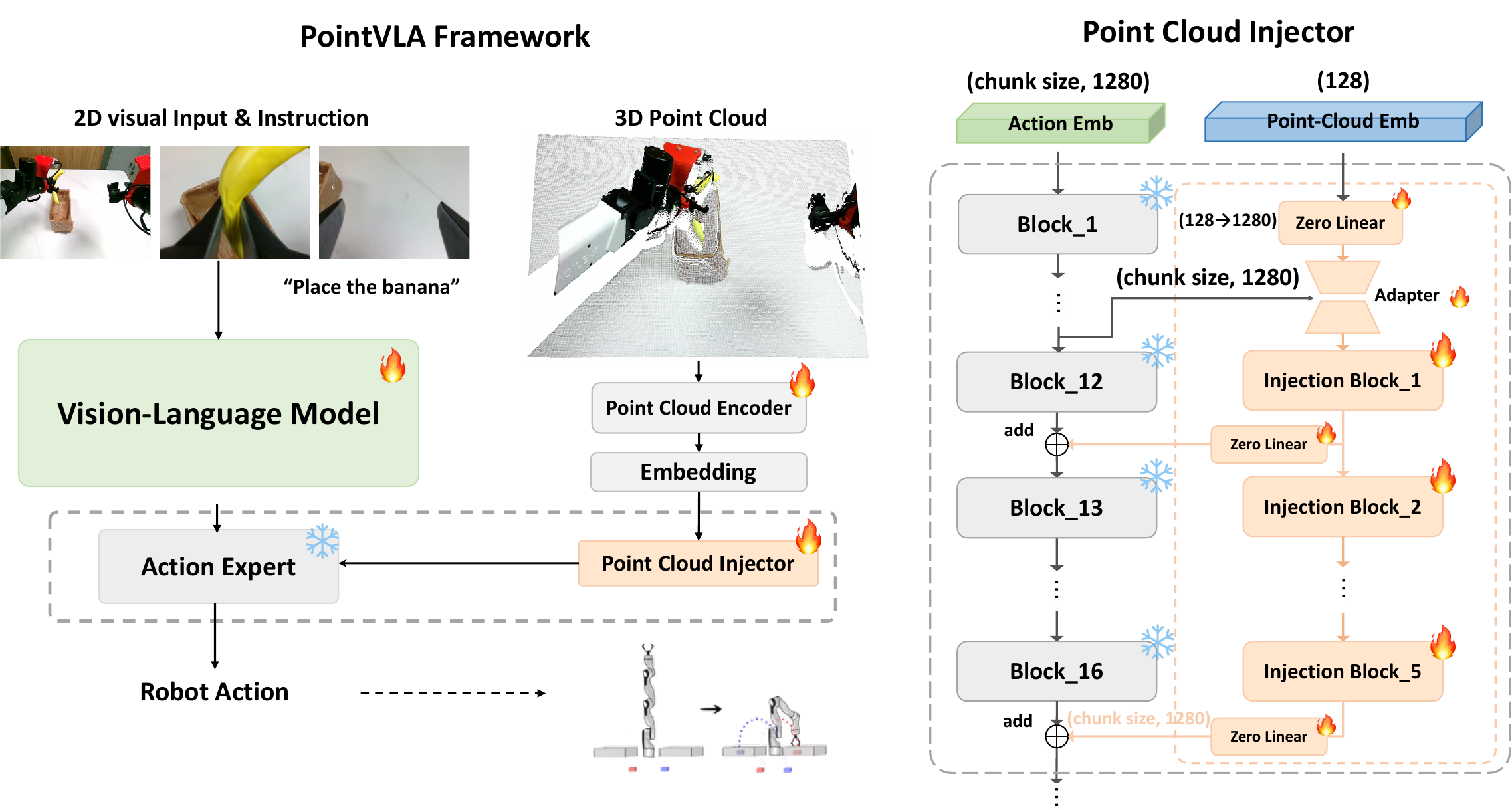}
    \caption{\textbf{Overview of PointVLA framework.} \textbf{Left:} The 2D image observation and instruction are processed by the vision-language model. The vanilla action expert remains frozen, while the new point cloud representation is integrated into the action expert through a modular network. \textbf{Right:} Details of the point cloud injector.}\label{fig:overall_framework}
\end{figure*}

Vision-Language-Action (VLA) models are driving a significant shift in real-world robot learning. Their power originates from the underlying Vision-Language Model (VLM), a robust backbone trained on vast internet datasets. This training enables effective alignment of image and text representations within a shared embedding space. The VLM acts as the model's 'brain,' processing instructions and current visual input to understand the task state. Subsequently, an 'action expert' module translates the VLM's state information into robot actions. This work builds upon DexVLA~\cite{wen2025dexvla}, which employs a 2 billion parameter Qwen2-VL~\cite{bai2025qwen25vl,wang2024qwen2vl} VLM as its backbone and a 1 billion parameter ScaleDP~\cite{scaledp} (a diffusion policy variant) as its action expert. DexVLA undergoes three training stages: a 100-hour cross-embodiment training phase (Stage 1) followed by embodiment-specific training (Stage 2), and an optional task-specific training (Stage 3) for complex tasks. All three stages utilize 2D visual input. While these VLA models exhibit impressive capabilities in diverse manipulation tasks, their reliance on 2D vision limits their performance in tasks requiring 3D understanding, such as object deception via photographs or generalization across varying table heights. The following section aim to illustrate how to inject the 3D world into the pre-trained VLA. The overall framework can be viewed at Figure~\ref{fig:overall_framework}.

\subsection{Injecting Point Cloud into VLA}
\textbf{Motivation.} As previously established, Vision-Language-Action (VLA) models are typically pre-trained on large-scale 2D robotic datasets. A critical observation underpinning our approach is the inherent disparity in data scale between existing 2D pretraining corpora and emerging 3D robotic datasets. Specifically, we posit that 3D sensor data (e.g., point clouds, depth maps) remains orders of magnitude smaller in volume compared to 2D vision-language datasets, a consequence of the extensive historical focus on 2D perception in robotics research. This discrepancy necessitates a methodology that preserves the rich visual representations learned from 2D pretraining while effectively integrating sparse 3D data.

A naive strategy to address this challenge involves directly converting 3D visual input into 3D visual tokens and blending them into the Large Language Model (LLM) - a popular approach that has been leveraged by many 3DVLM, such as LLaVA-3D~\cite{zhu2024llava3d}. However, current vision-language models exhibit limited 3D comprehension when fine-tuned on small-scale 3D datasets, a limitation exacerbated by two factors: (1) the substantial domain gap between 2D pixels and 3D geometric structures, and (2) the scarcity of high-quality 3D-text paired data compared to the abundance of image-text and text-only corpora. To circumvent these issues, we propose a paradigm that treats 3D point cloud data as a complementary conditioning signal rather than a primary input modality. This strategy decouples 3D processing from the core 2D visual encoder, thereby preserving the integrity of pretrained 2D representations while enabling the model to leverage geometric cues. By design, our approach mitigates catastrophic forgetting of 2D knowledge and reduces the risk of overfitting to limited 3D data.

\textbf{Model architecture for point cloud injector.} The overall architecture of the point cloud injector is shown in Figure~\ref{fig:framework} (right). Specifically, for the incoming point cloud embedding, we first transform the channel dimension to match that of the vanilla action expert. Since the action embedding from the point cloud can be large, depending on the chunk size, we design an action embedding bottleneck to compress the information from the action expert while aligning it with the 3D point cloud embedding. For selected blocks in the action expert, we first apply an MLP layer as an adapter for each block, followed by an addition operation to inject the point cloud embedding into the model. 

Notably, we avoided injecting the 3D features into every block of the action expert for two primary reasons. First, the computational cost would be prohibitively high due to the required conditioning blocks. Second, injection inevitably alters the model's representation at the affected block. Given that we aimed to minimize interference from the limited 3D visual knowledge on the pre-trained action embedding derived from 2D visual input, we conducted an analysis to identify blocks that could be skipped during inference without compromising performance. We subsequently injected 3D features only into these less critical blocks.

\textbf{Point cloud encoder.} Consistent with observations in DP3~\cite{ze20243d} and iDP3~\cite{ze2024generalizable}, we found that pre-trained 3D visual encoders hindered performance, often preventing successful robot behavior learning in new environments. Therefore, we adopted a simplified, hierarchical convolutional architecture. The upper convolutional layers extracted low-level features, while the lower convolutional blocks learned high-level scene representations. The max pooling was employed between layers to progressively reduce point cloud density. Finally, we concatenated feature embeddings from each convolutional block into a unified embedding, encapsulating multi-level 3D representation knowledge. The extracted point cloud feature embedding is retained for subsequent use. This architecture is similar to the iDP3 encoder. We believe that employing a more advanced point cloud encoder could further enhance model performance. However, as this is not the core novelty of our approach, we leave it for future discussion.

\begin{figure}[t]
    \centering
    \includegraphics[width=0.4\textwidth]{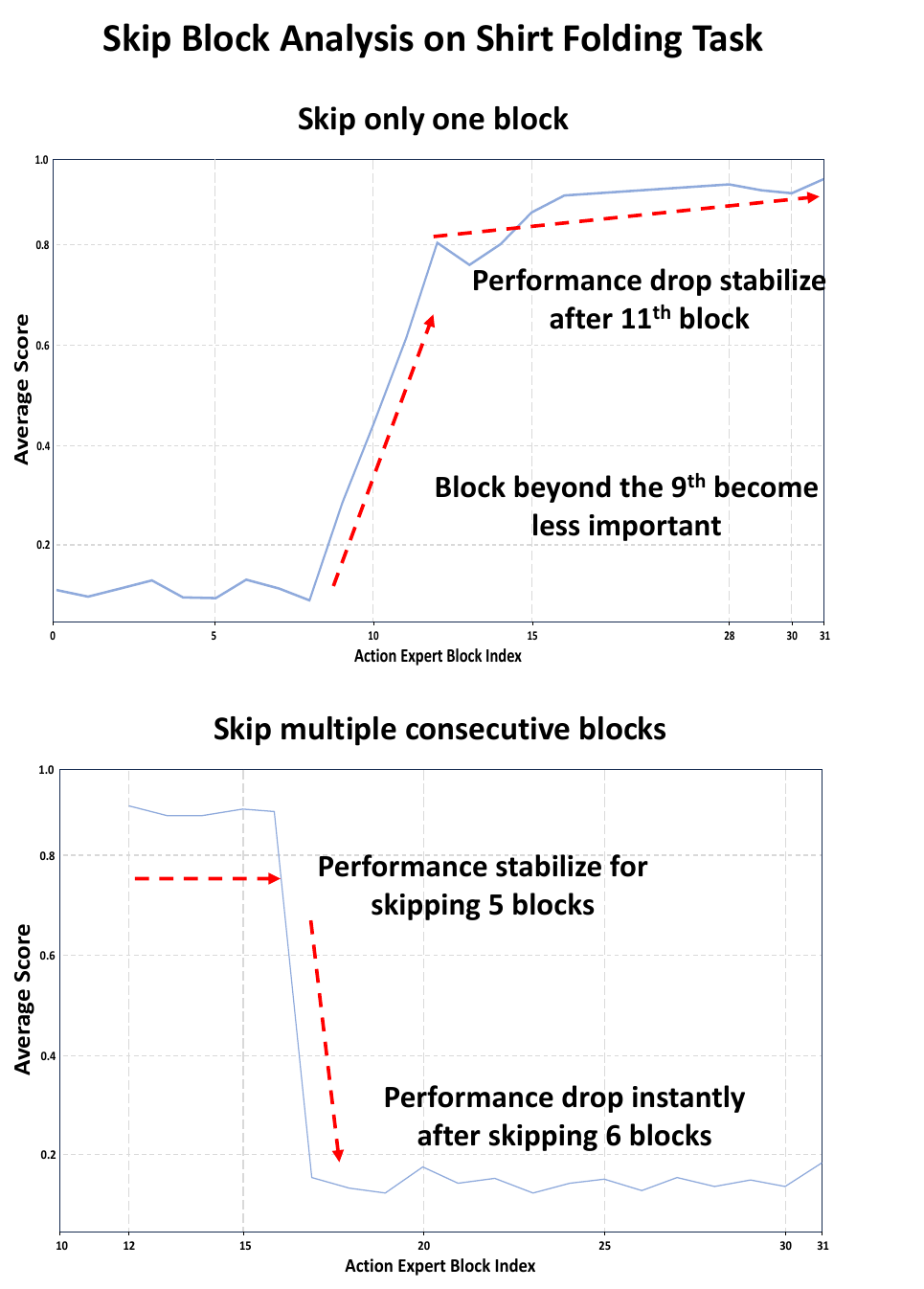}
    \caption{\textbf{Skip block analysis for action expert in VLA model.} \textbf{Left:} skipping only one block at a time. \textbf{Right:} Skipping multiple consecutive blocks starting from the 11th block.}\label{fig:skip_block}
\end{figure}

\subsection{Which Blocks to Inject Point Cloud? A Skip Block Analysis}
As mentioned earlier, injecting the point cloud into every block of the action expert is not ideal, as it increases computational cost and disrupts the vanilla action representation learned from extensive 2D visual-based robot data. Therefore, we analyze which blocks in the action expert are less critical—i.e., those that can be skipped during inference without affecting performance. This approach aligns conceptually with techniques used in image generation, vision models, and large language models~\cite{jaiswal2024ffn, shen2022sliced, flux2024, flux1-lite}. Specifically, we use shirt folding in DexVLA~\cite{wen2025dexvla} as a case study for our analysis. Recall that the DexVLA is equipped with 1 billion parameter action expert with 32 diffusion transformer blocks. The evaluation follows the same metrics—average score, a standard measure for long-horizon tasks~\cite{[pi0, wen2025dexvla, mees2022calvin}—by dividing the task into multiple steps and assessing performance based on step completion. We start by skipping one block at a time and summarize our findings in the figure below.

We illustrated the results in Figure~\ref{fig:skip_block} (left). Our experiment reveals that the first 11 blocks are crucial for the model—skipping any of them leads to a significant drop in performance. Specifically, when blocks before layer 11 are skipped, the gripper fails to close tightly, making it difficult for the model to complete the task. However, starting from the 11 block onward, skipping a single block becomes acceptable, up until the last block. This indicates that blocks 11 to 31 contribute less to performance after training. To further investigate which blocks would be fit for point cloud injection, we conduct a multi-block skipping analysis from block 11 onward, as shown in Figure~\ref{fig:skip_block} (right). We find that up to five consecutive blocks can be skipped before the model fails at task completion. This suggests that 3D representation can be selectively injected into the action expert through specific blocks, optimizing efficiency without significantly impacting performance. As a result, we set all 3D conditioning blocks to be trainable when new data is introduced. We freeze all modules in the vanilla action expert except for the final layers, which are adjusted to fit the embodiment’s output. Ultimately, we only train five additional injection blocks, which are lightweight and fast during inference, making our approach highly cost-efficient.

\begin{figure*}[t]
    \centering
    \includegraphics[width=0.97\textwidth]{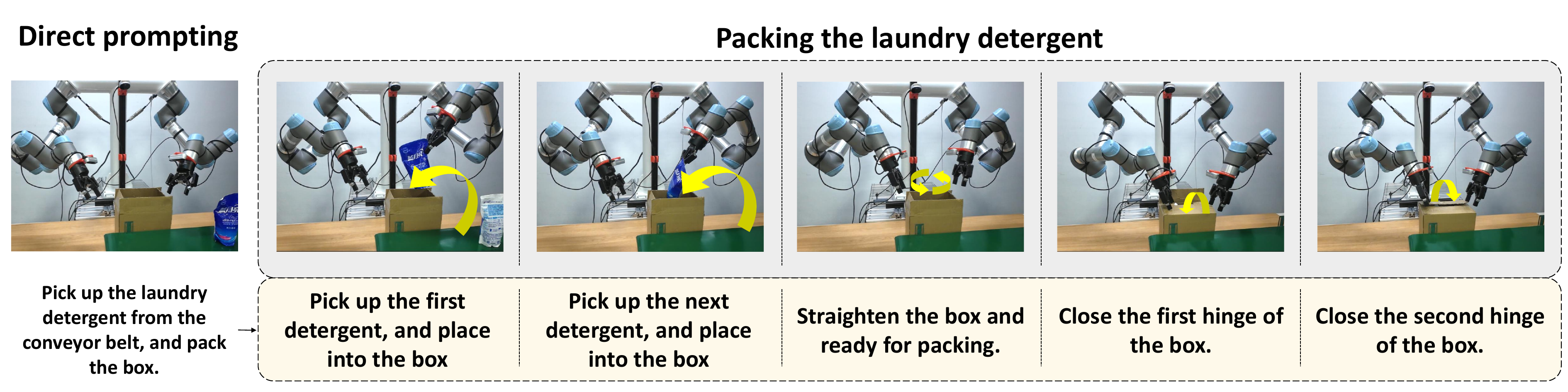}
    \caption{\textbf{Setup for bimanual UR5e.} We utilize three cameras: two RealSense D435i mounted on the wrists and one RealSense L515 positioned above. Our model is evaluated on a challenging long-horizon task that involves picking up two laundry detergent bottles from a moving conveyor belt and packing them into a box.}\label{fig:long_horizon_example}

\end{figure*}

\begin{figure*}[t]
    \centering
    \includegraphics[width=0.85\textwidth]{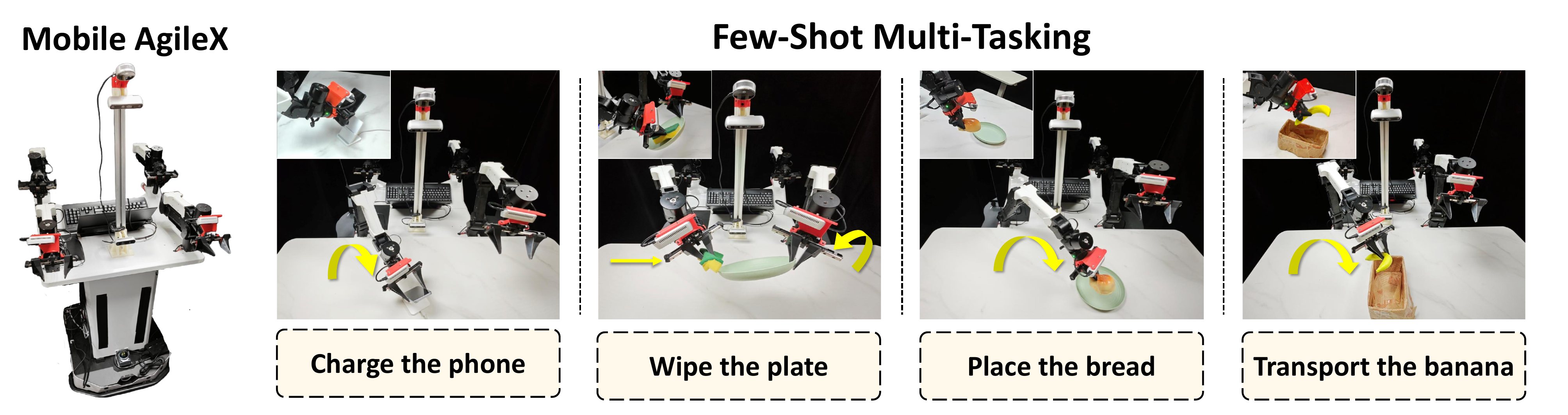}
    \caption{\textbf{Setup for bimanual AgileX.} We utilize three cameras: two RealSense D435i mounted on the wrists and one RealSense L515 positioned above. Our model is evaluated on four tasks in a few-shot setting.}\label{fig:multi_task_example}

\end{figure*}

\section{Experiment}
This section presents a series of experiments to validate the effectiveness of our approach. Specifically, we provide detailed descriptions of the real robot and workspace setup in Section~\ref{sec:implment_details}. We evaluate the few-shot multi-tasking setting and long-horizon challenging tasks in Sections~\ref{sec:fewshot} and~\ref{sec:longhorizon}, respectively. In Sections~\ref{sec:screen} and~\ref{sec:height}, we explore two unique types of generalization that can only be observed with 3D visual input. Finally, we compare our method against simulation benchmarks. 

\begin{figure}[t]
    \centering
    \includegraphics[width=0.5\textwidth]{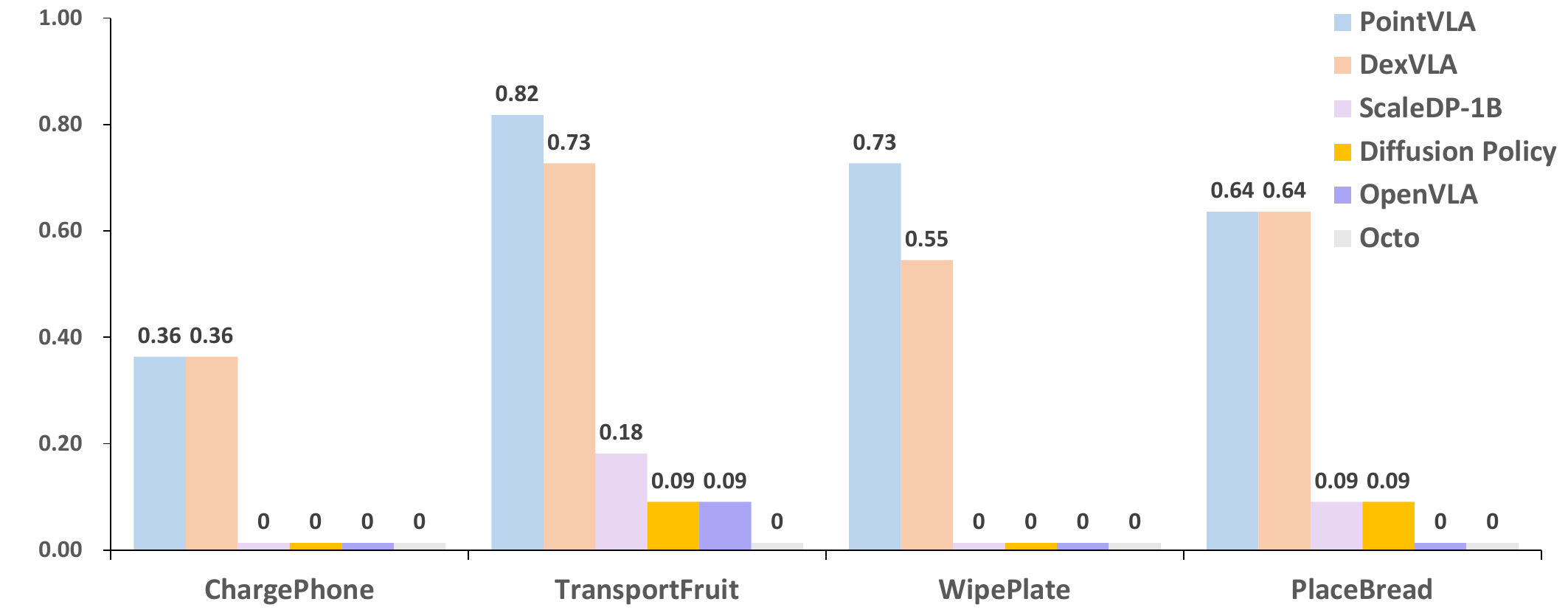}
    \caption{\textbf{Experimental results on few-shot multi-tasking on bimanual AgileX.}}\label{fig:multitask_experiment}
\end{figure}

\subsection{Implementation Details}
\label{sec:implment_details}
In this work, we conduct real robot experiments on two embodiments:
\begin{itemize}
    \item \textbf{Bimanual UR5e}. Two UR5e robots, each with a Robotiq parallel jaw gripper and a wrist-mounted camera. A top-down camera is positioned between the two arms. This setup has a total of three camera views and a 14-dimensional configuration and action space. Data is collected at 15Hz. We use the RealSense D435i camera as the wrist camera. 
    \item \textbf{Bimanual AgileX}. Two 6-DoF AgileX arms, each with a wrist-mounted camera and a base camera. This setup has a 14-dimensional configuration and action space, supported by three cameras in total. Data is collected at 30Hz. We use the RealSense D435i camera as a wrist camera. 
\end{itemize}
We use RealSense L515 camera to collect the point cloud. We set the VLM model trainable as the model needs to learn new language instruction. For both experiments, we use stage 1 pre-trained weights from DexVLA~\cite{wen2025dexvla} and fine-tune for our model. We use the same training hyper-parameters as the stage 2 training in DexVLA, and use the last checkpoint for evaluation to avoid cherry picking. We set chunk size to 50 for all tasks. 

\textbf{Baseline.} In our experiments, we compared with many state-of-the-art model, including the Diffusion Policy (DP)~\cite{diffusion-policy}, 3D Diffusion Policy (DP3)~\cite{ze20243d}, ScaleDP-1B~\cite{scaledp}, a variant of scaling up diffusion policy to 1B parameters, Octo~\cite{octo}, OpenVLA~\cite{openvla}, and DexVLA~\cite{wen2025dexvla}. Note that since PointVLA is built on top of DexVLA, the DexVLA can be viewed as an ablation of our proposed PointVLA without the incorporation of 3D point cloud data. 

\begin{figure}[t]
    \centering
    \includegraphics[width=0.48\textwidth]{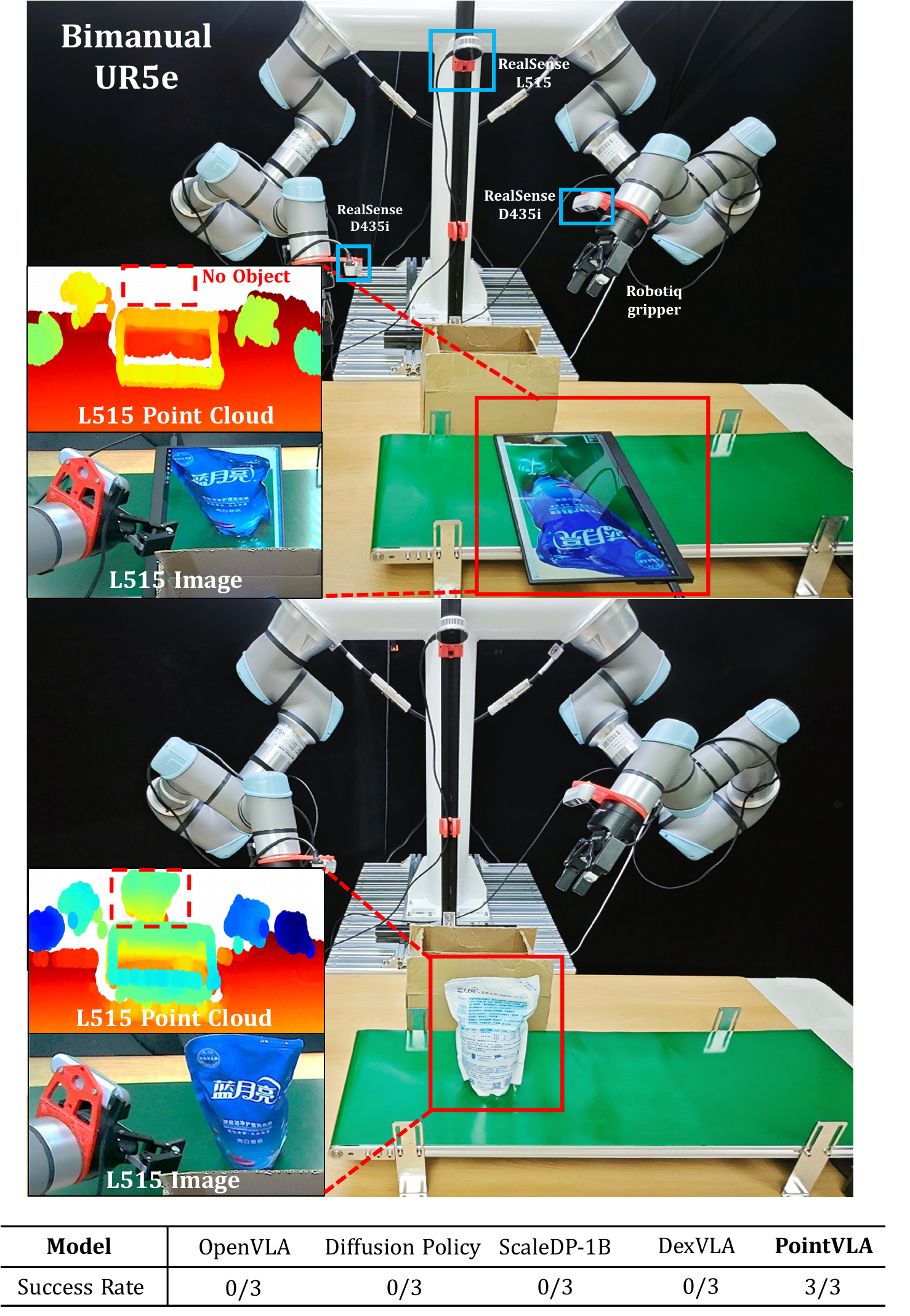}
    \caption{\textbf{The experimental setup for real-vs-photo discrimination.} We replace the real laundry detergent with its photo displayed on a screen placed on the conveyor belt. From an exocentric view, the image appears noticeably different from the real object. However, from the egocentric top camera’s perspective, the photo closely resembles the actual detergent. We show experimental results on the bottom table.}\label{fig:photovsreal}
\end{figure}

\subsection{Few-Shot Multi-Tasking}
\label{sec:fewshot}
\textbf{Task description.} As illustrated in Fig~\ref{fig:multi_task_example}, we designed four few-shot tasks for our real-world experiment: \textbf{ChargePhone}, \textbf{WipePlate}, \textbf{PlaceBread}, \textbf{TransportFruit}. Objects were placed randomly within a small range, and we report the average success rate for each method. \textbf{1) ChargePhone}: The robot picks up a smartphone and places it on a wireless charger. The phone's size tests action precision, while its fragility requires careful handling. \textbf{2) WipePlate}: The robot simultaneously picks up a sponge and a plate, using the sponge to wipe the plate, assessing bimanual manipulation skills. \textbf{3) PlaceBread}: The robot picks up a piece of bread and places it on a plate. A thin foam layer under the bread ensures height generalization testing. \textbf{4) TransportFruit}: The robot picks up a randomly oriented banana and places it in a centrally located box.

As our approach aims to verify the few-shot multi-tasking ability of the model, we collected 20 demonstrations for each task, resulting in a total of 80 demonstrations. The objects' positions are randomized within a small space. These tasks assess the model's capability to manage both independent and coordinated robot movements across diverse scenarios. All data are collected using 30Hz. 

\textbf{Experimental results.} The experimental results are presented in Table~\ref{fig:multitask_experiment}, where our method outperforms all baselines in this scenario. Notably, the Diffusion Policy fails in most cases, likely because the sample size for each task is too small, causing the action representation space to become entangled—an observation consistent with previous findings in the previous literature~\cite{wu2024discrete}. Furthermore, even increasing the model size (ScaleDP-1B) does not lead to significant improvement.

DexVLA demonstrates strong few-shot learning capabilities despite the limited data; however, its performance remains on par with or inferior to PointVLA. The incorporation of point cloud data in PointVLA enables more sample-efficient learning, emphasizing the necessity of integrating 3D information into the model. More importantly, our results confirm that our method successfully preserves the ability to learn from the 2D pre-trained VLA.

\subsection{Long-Horizon Task: Packing on Assembly Line}
\label{sec:longhorizon}
Beyond conventional multi-tasking, we further fine-tune PointVLA on long-horizon packing tasks, as illustrated in Figure \ref{fig:long_horizon_example}. This is an exceptionally challenging task for several reasons. First, the assembly line is in motion, requiring the robot to quickly and precisely grasp objects. Second, the embodiment in this scenario differs from those present in the pre-trained data, necessitating rapid adaptation to a completely new setup. Third, as a long-range task, the robot must sequentially pick and place two bags of laundry detergent before sealing the packing box. These complexities make the task highly demanding.

The evaluation metrics are detailed in the Appendix. As shown in Table \ref{tbl:taskscore_longhorizon}, PointVLA achieves the highest average length in long-range tasks, surpassing DexVLA, a strong baseline, by 0.64. It also outperforms several other baselines. However, the next section highlights an even more critical aspect of PointVLA, the object hallucination problem.

\input{tables/long_horizon}

\subsection{Real-vs-Photo Discrimination}
\label{sec:screen}
In this section, we explore a unique setup called real-vs-photo discrimination. Specifically, we replace the real object with a picture of the object. From a 2D perspective, the "fake" object displayed on a screen appears nearly identical to the real object, yet it does not physically exist. Humans can easily recognize this discrepancy and refrain from reaching for the object because we understand it is not real—but can a robot model do the same?

To illustrate this scenario, we conduct an experiment using a bimanual UR5e in a packing task. We set the experiments by replacing laundry detergent with a photo of the laundry detergent projected on a screen. The experimental setup can be found in Figure~\ref{fig:photovsreal}. From an exocentric view,
the image appears noticeably different from the real object. How-
ever, from the egocentric top camera’s perspective, the photo closely
resembles the actual detergent. We observe that conventional 2D-based vision-language-action models, such as OpenVLA and DexVLA, fail to distinguish between the image and a real object. These models attempt to grasp the object, and in the case of DexVLA, it repeatedly tries to pick up the non-existent detergent. Since the model believes the object is present but continuously fails to grasp it, it enters a repetitive grasping loop. In contrast, PointVLA successfully recognizes that no real object exists on the conveyor belt. By leveraging 3D spatial understanding, it determines that the space where the object should be is actually empty. This key advantage highlights the strength of our approach, demonstrating the superiority of 3D-aware models in mitigating object hallucination.

\subsection{Height Adaptability}
\label{sec:height}
In our context, height generalization refers to a model's ability to adapt to different table heights. This is a crucial capability for robotic models, as most demonstrations are conducted on a fixed table height. However, what happens when the robot is deployed in an environment with a significantly different table height than what it was trained on?

\begin{figure}[t]
    \centering
    \includegraphics[width=0.48\textwidth]{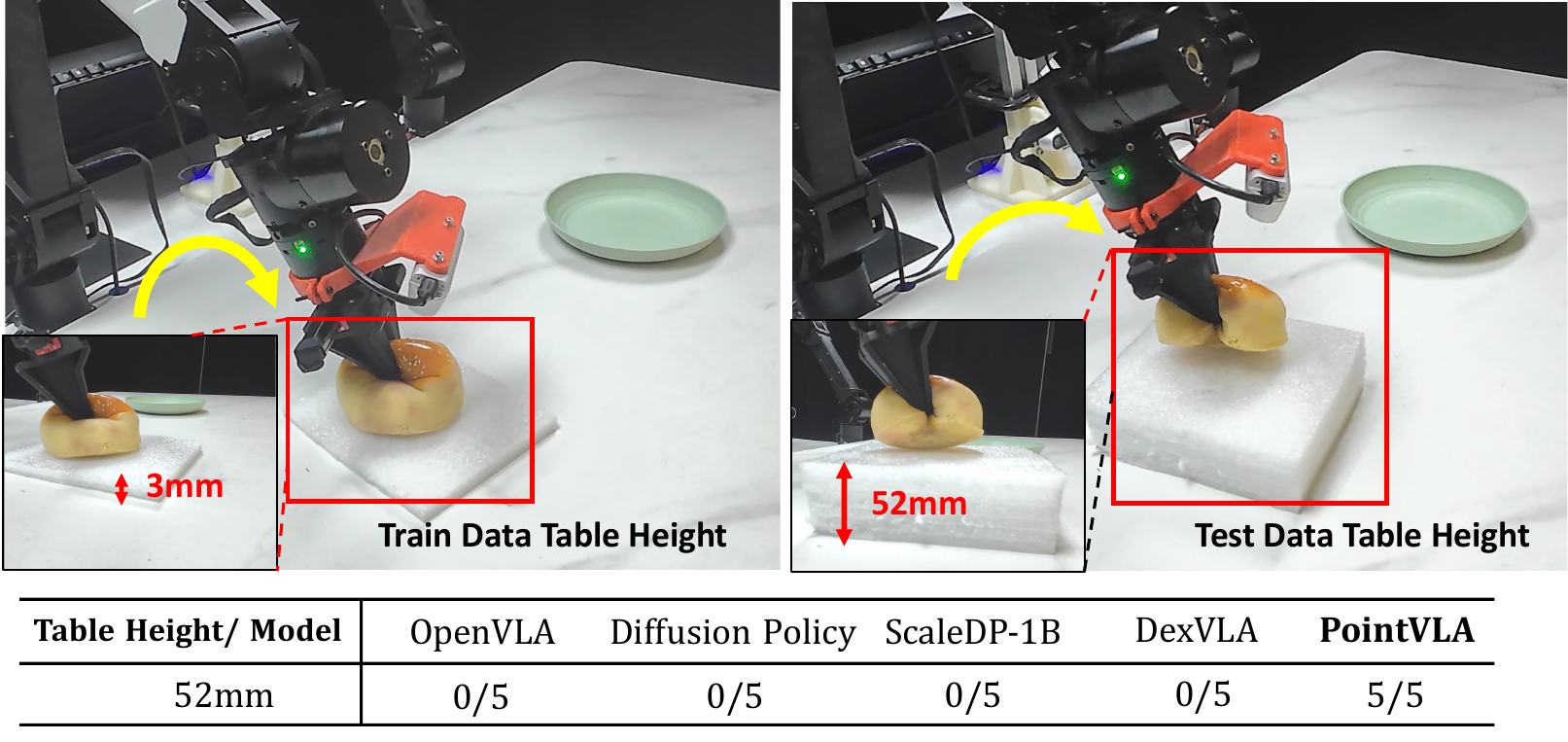}
    \caption{\textbf{Height adaptability for PointVLA}. Even if the model is trained on a standard height, when PointVLA encounters different table heights at test time, it can adapt to the new height and successfully complete the task. In contrast, conventional 2D vision-based imitation learning methods fail completely.}\label{fig:table_height}
\end{figure}

To investigate this question, we designed an experiment illustrated in Figure~\ref{fig:table_height}. Specifically, in the "place bread" task, we placed a foam plastic layer beneath the bread. During training, this foam layer was 3mm thick, and all collected data was based on this height. During inference, we increased the foam thickness to 52mm to evaluate the model’s height generalization. Our observations show that conventional 2D-based VLA models, such as OpenVLA~\cite{openvla}, DP~\cite{diffusion-policy}, ScaleDP-1B~\cite{scaledp}, and DexVLA~\cite{wen2025dexvla} all failed in this scenario. Upon detecting the bread, these models attempted to push it down and grasp it at the height seen in the training data, failing to adjust to the increased height. In contrast, PointVLA successfully completed the task. By leveraging the point cloud data, it accurately perceived the bread’s new height, adjusted its gripper accordingly, and executed a successful pick-up. This experiment demonstrates that incorporating 3D information enables VLA models to handle variations in object height—an ability that purely 2D-based models lack.

\subsection{Experimental Results on Simulation Benchmarks}
\label{sec:simulation}
We evaluate our approach on RoboTwin~\cite{mu2024robotwin}, a widely used mobile bimanual platform equipped with 14-degree-of-freedom robots. This benchmark encompasses a diverse set of tasks. We compare our method against Diffusion Policy~\cite{diffusion-policy} and 3D Diffusion Policy (DP3)~\cite{ze20243d}. Diffusion Policy is a well-established baseline for visuomotor policy learning, while DP3 extends it to the 3D domain. The vanilla DP3 uses only point cloud data as input. For a fair comparison, we also incorporate RGB images into DP3. The implementation is carried out by RoboTwin. We compare both versions of DP3 alongside the vanilla Diffusion Policy. In all experiments, we use a standard image resolution of $320 \times 180$ for camera input, including the L515 and the top camera.

The tests were conducted using datasets of 20 and 50 samples. Following the training setup in RoboTwin, the policy was trained using three random seeds (0, 1, 2) without cherry picking for each experiment. Each policy was then tested 100 times, yielding three success rates. The mean and standard deviation of these success rates were computed to obtain the experimental results presented below. 

The experimental results are presented in Table~\ref{tab:robotwin}. Baseline results, including those for the 3D Diffusion Policy and Diffusion Policy, are reported by RoboTwin. Notably, across all tasks and diverse settings, our proposed PointVLA achieves the highest average success rate, regardless of whether it is trained on 20 or 50 demonstrations. This demonstrates that our method is effective even when data resources are limited and continues to perform well when abundant training data is available.

Additionally, we observe that for purely 3D models like DP3, directly incorporating RGB input can negatively impact performance. In contrast, our approach highlights the necessity of conditionally integrating 3D point cloud data into the model, which significantly enhances performance compared to models that rely solely on 2D visual input.

\input{tables/robotwin}

\section{Conclusion}
While Vision-Language-Action (VLA) models excel in robot learning through large-scale 2D pretraining, their reliance on RGB inputs limits 3D spatial reasoning. Retraining with 3D data is costly, and discarding 2D datasets reduces generalization. To address this, we introduced PointVLA, a framework that enhances pre-trained VLAs with 3D point cloud inputs while preserving their 2D representations. By integrating a modular 3D feature injector and leveraging skip block analysis, our method incorporates spatial information efficiently without full retraining. Experiments in both simulated and real-world environments demonstrate PointVLA’s effectiveness, achieving few-shot multi-task learning (4 tasks with only 20 demonstrations each) and excelling in long-horizon tasks like dynamic item packing. Real-world tests on bimanual robots (UR5e and AgileX arms) further validate its practicality and safety. Our work highlights the feasibility of augmenting pre-trained robot models with new modalities without costly retraining. Future work includes expanding 3D-aware pretraining across larger datasets.

{
    \small
    \bibliographystyle{ieeenat_fullname}
    \bibliography{main}
}

\end{document}

%% file: tables/long_horizon.tex
\begin{table}[tb]
  \centering
  \caption{\textbf{Experimental results for long-horizon task on bimanual UR5e.} \textit{The task is completed in a sequence.} The Avg. Len. denotes the average success length of the model.}

  \label{tbl:taskscore_longhorizon}
  \resizebox{\linewidth}{!}{
      \begin{tabular}{c|ccccc|c}
        \toprule
        \multirow{2}{*}{Method} & \multicolumn{5}{c|}{Packing Laundry Detergent} \\
        & 1 & 2 & 3 & 4 & 5 & Avg. Len. \\ 
        \midrule
    Octo~\cite{octo} & 1/11 & 1/11 & 0 & 0 & 0 & 0.27 \\
    OpenVLA~\cite{kim24openvla} & 2/11 & 1/11 & 0 & 0 & 0 & 0.36  \\
    Diffusion Policy~\cite{diffusion-policy} & 2/11 & 1/11 & 0 & 0 & 0 & 0.36 \\
    ScaleDP-1B~\cite{scaledp} & 4/11 & 2/11 & 0 & 0 & 0 & 0.72 \\
    DexVLA~\cite{wen2025dexvla} & 2/11 & 5/11 & 1/11 & 1/11 & 0 & 1.72 \\
   \textbf{PointVLA} & 3/11 & 1/11 & 1/11 & 2/11 & 2/11 & 2.36  \\
        \bottomrule
      \end{tabular}
}
\end{table}

%% file: tables/robotwin.tex
\begin{table}[t]
    \centering
    \scriptsize
    \resizebox{0.5\textwidth}{!}{\begin{tabular}{l cc | l cc}
        \hline
        \multirow{2}{*}{\textbf{Task}} & \multicolumn{2}{c|}{\textbf{Number of Demonstrations}} & \multirow{2}{*}{\textbf{Task}} & \multicolumn{2}{c}{\textbf{Number of Demonstrations}} \\
         & 20 & 50 &  & 20 & 50 \\
        \midrule
        \textit{Block Hammer Beat} &  &  & \textit{Block Handover} &  &  \\
        \midrule
        DP3 (Point Cloud) & 47.7 $\pm$ 7.4 & 58.3 $\pm$ 6.5 & DP3 (Point Cloud) & 82.7 $\pm$ 6.1 & 85.0 $\pm$ 15.6 \\
        DP3 (Point Cloud+RGB) & 44.7 $\pm$ 3.8 & 79.0 $\pm$ 2.0 & DP3 (Point Cloud+RGB) & 88.7 $\pm$ 5.0 & 94.3 $\pm$ 7.2 \\
        DP & 0.0 $\pm$ 0.0 & 0.0 $\pm$ 0.0 & DP & 0.0 $\pm$ 0.0 & 0.0 $\pm$ 0.0 \\
        PointVLA & \textbf{61.2 $\pm$ 4.8} & \textbf{84.6 $\pm$ 3.9} & PointVLA & \textbf{89.7 $\pm$ 4.3} & \textbf{96.1 $\pm$ 1.9} \\
        \midrule
        \textit{Blocks Stack (Easy)} &  &  & \textit{Blocks Stack (Hard)} &  &  \\
        \midrule
        DP3 (Point Cloud) & 3.3 $\pm$ 3.2 & 17.0 $\pm$ 7.0 & DP3 (Point Cloud) & 0.3 $\pm$ 0.6 & 1.7 $\pm$ 0.6 \\
        DP3 (Point Cloud+RGB) & 2.7 $\pm$ 1.2 & 17.0 $\pm$ 1.0 & DP3 (Point Cloud+RGB) & 0.0 $\pm$ 0.0 & 2.0 $\pm$ 2.0 \\
        DP & 0.0 $\pm$ 0.0 & 0.0 $\pm$ 0.0 & DP & 0.0 $\pm$ 0.0 & 0.0 $\pm$ 0.0 \\
        PointVLA & \textbf{10.8 $\pm$ 2.8} & \textbf{24.3 $\pm$ 4.7} & PointVLA & \textbf{2.3 $\pm$ 0.9} & \textbf{3.7 $\pm$ 1.4} \\
        \midrule
        \textit{Bottle Adjust} &  &  & \textit{Container Place} &  &  \\
        \midrule
        DP3 (Point Cloud) & 55.7 $\pm$ 1.5 & 70.7 $\pm$ 2.5 & DP3 (Point Cloud) & 52.7 $\pm$ 4.5 & 74.0 $\pm$ 5.6 \\
        DP3 (Point Cloud+RGB) & 28.3 $\pm$ 12.9 & 27.7 $\pm$ 16.5 & DP3 (Point Cloud+RGB) & 38.0 $\pm$ 7.9 & 58.3 $\pm$ 5.9 \\
        DP & 13.0 $\pm$ 11.8 & 24.7 $\pm$ 13.8 & DP & 5.3 $\pm$ 4.2 & 16.3 $\pm$ 2.5 \\
        PointVLA & \textbf{61.2 $\pm$ 2.8} & \textbf{74.5 $\pm$ 9.7} & PointVLA & \textbf{58.5 $\pm$ 4.8} & \textbf{81.3 $\pm$ 3.9} \\
        \midrule
        \textit{Diverse Bottles Pick} &  &  & \textit{Dual Bottles Pick (Easy)} &  &  \\
        \midrule
        DP3 (Point Cloud) & 13.3 $\pm$ 5.5 & 34.7 $\pm$ 6.7 & DP3 (Point Cloud) & 37.0 $\pm$ 4.6 & 60.3 $\pm$ 7.1 \\
        DP3 (Point Cloud+RGB) & 0.7 $\pm$ 0.6 & 5.3 $\pm$ 2.1 & DP3 (Point Cloud+RGB) & 29.7 $\pm$ 3.5 & 67.3 $\pm$ 9.3 \\
        DP & 0.0 $\pm$ 0.0 & 0.3 $\pm$ 0.6 & DP & 1.3 $\pm$ 1.5 & 26.7 $\pm$ 3.1 \\
        PointVLA & \textbf{21.5 $\pm$ 3.8} & \textbf{38.7 $\pm$ 5.7} & PointVLA & \textbf{41.2 $\pm$ 3.2} & \textbf{68.5 $\pm$ 4.8} \\
        \midrule
        \textit{Dual Bottles Pick (Hard)} &  &  & \textit{Dual Shoes Place} &  &  \\
        \midrule
        DP3 (Point Cloud) & 33.0 $\pm$ 2.6 & 48.0 $\pm$ 5.2 & DP3 (Point Cloud) & 5.7 $\pm$ 0.6 & 10.0 $\pm$ 2.6 \\
        DP3 (Point Cloud+RGB) & 23.0 $\pm$ 2.0 & 46.3 $\pm$ 7.8 & DP3 (Point Cloud+RGB) & 1.7 $\pm$ 2.9 & 3.7 $\pm$ 0.6 \\
        DP & 2.0 $\pm$ 1.7 & 32.3 $\pm$ 5.9 & DP & 0.0 $\pm$ 0.0 & 3.0 $\pm$ 1.7 \\
        PointVLA & \textbf{34.2 $\pm$ 2.7} & \textbf{51.3 $\pm$ 3.2} & PointVLA & \textbf{7.5 $\pm$ 1.8} & \textbf{13.1 $\pm$ 2.9} \\
        \midrule
        \textit{Empty Cup Place} &  &  & \textit{Mug Hanging (Easy)} &  &  \\
        \midrule
        DP3 (Point Cloud) & 33.0 $\pm$ 6.2 & 70.3 $\pm$ 7.2 & DP3 (Point Cloud) & 7.3 $\pm$ 2.9 & 14.0 $\pm$ 3.6 \\
        DP3 (Point Cloud+RGB) & 26.3 $\pm$ 10.4 & 71.3 $\pm$ 4.0 & DP3 (Point Cloud+RGB) & 1.0 $\pm$ 1.0 & 2.0 $\pm$ 2.0 \\
        DP & 0.3 $\pm$ 0.6 & 14.7 $\pm$ 6.0 & DP & 0.0 $\pm$ 0.0 & 0.0 $\pm$ 0.0 \\
        PointVLA & \textbf{41.2 $\pm$ 4.8} & \textbf{73.5 $\pm$ 4.2} & PointVLA & \textbf{9.3 $\pm$ 1.9} & \textbf{19.1 $\pm$ 2.8} \\
        \midrule
        \textit{Mug Hanging (Hard)} &  &  & \textit{Pick Apple Messy} &  &  \\
        \midrule
        DP3 (Point Cloud) & 12.7 $\pm$ 0.6 & 11.0 $\pm$ 6.1 & DP3 (Point Cloud) & 5.7 $\pm$ 4.5 & 10.7 $\pm$ 4.0 \\
        DP3 (Point Cloud+RGB) & 0.0 $\pm$ 0.0 & 2.0 $\pm$ 2.0 & DP3 (Point Cloud+RGB) & 6.7 $\pm$ 2.3 & 28.7 $\pm$ 9.5 \\
        DP & 0.0 $\pm$ 0.0 & 0.3 $\pm$ 0.6 & DP & 3.3 $\pm$ 1.5 & 6.0 $\pm$ 5.0 \\
        PointVLA & \textbf{13.5 $\pm$ 1.7} & \textbf{16.2 $\pm$ 1.9} & PointVLA & \textbf{9.3 $\pm$ 2.8} & \textbf{31.5 $\pm$ 4.7} \\
        \hline
    \end{tabular}}
    \caption{\textbf{Experimental results on RoboTwin~\cite{mu2024robotwin}}. Performance comparison across different tasks and demonstrations (results for 20 and 50 demonstrations).}
    \label{tab:robotwin}
\end{table}